\pdfoutput=1

\documentclass[11pt]{article}

\usepackage{EMNLP2023}

\usepackage{times}
\usepackage{latexsym}
\usepackage{xcolor}
\usepackage{colortbl}
\usepackage[T1]{fontenc}

\usepackage[utf8]{inputenc}

\usepackage{microtype}

\usepackage{inconsolata}

\usepackage{graphicx}

\usepackage{xcolor}
\usepackage{lipsum}
\usepackage{booktabs}
\usepackage{multicol}
\usepackage{multirow}
\usepackage{graphicx}
\usepackage{tabularx}
\usepackage{float}
\usepackage{bm}
\usepackage{todonotes}
\usepackage{xargs}
\usepackage{xspace}
\usepackage{amsmath}
\usepackage{cleveref}
\usepackage{subcaption}
\usepackage{mathtools}
\usepackage{bigstrut}
\usepackage{tikz}
\usetikzlibrary{shapes}
\usepackage{rotating}
\usepackage{siunitx}
\usepackage{amssymb}
\usepackage{arydshln}
\usepackage{listings}
\usepackage{xfrac}

\usepackage{caption}

\usepackage{dcolumn}
\newcolumntype{L}{D{.}{.}{2,1}}

\setlipsum{%
  par-before = \begingroup\color{gray},
  par-after = \endgroup
}
\usepackage[inline]{enumitem}
\usepackage{tabularx}

\newcommand*{\ie}{i.e.\@\xspace}

\makeatletter\newcommand*{\etc}{%
	\@ifnextchar{.}%
	{etc}%
	{etc.\@\xspace}%
}
\makeatother

\newcommand{\lde}{\textsc{De}\@\xspace}
\newcommand{\len}{\textsc{En}\@\xspace}
\newcommand{\lro}{\textsc{Ro}\@\xspace}

\newcommand{\adam}{\textsc{Adam}\@\xspace}
\newcommand{\sacrebleu}{\texttt{sacreBLEU}\xspace}
\newcommand{\chrf}{\textsc{chrF}\xspace}
\newcommand{\comet}{\textsc{Comet}\xspace}
\newcommand{\spBLEU}{\textsc{spBLEU}\@\xspace}
\newcommand{\bleu}{\textsc{Bleu}\@\xspace}
\newcommand{\knn}{k\text{-NN}}
\newcommand{\cka}{\textsc{cka}\@\xspace}
\newcommand{\lns}{\textsc{lns}\@\xspace}

\definecolor{red}{RGB}{141,45,57}
\definecolor{dark}{RGB}{55,65,74}
\definecolor{blue}{RGB}{0,105,170}
\definecolor{gold}{RGB}{174,159,109}
\definecolor{gray}{RGB}{175,179,183}
\definecolor{darkgreen}{RGB}{50,110,30}

\definecolor{ptgreen}{RGB}{213,232,212}

\makeatletter
\newcommand*\standardbin{+}
\newcommand*\tabularbin[1]{%
  \mathbin{\mathpalette{\@tabularsym\standardbin}{#1}}%
}
\newcommand*\@tabularsym[3]{%
  \setbox\z@\hbox{$#2#1\m@th$}%
  \hbox to\wd\z@{\hss$#2#3\m@th$\hss}%
}
\makeatother

\newcolumntype{Y}{>{\centering\arraybackslash}X}

\newcommandx{\hendra}[2][1=]{\vspace{0.2cm} \todo[linecolor=blue,backgroundcolor=blue!15,bordercolor=blue, #1]{\textbf{Hendra:} #2}}
\newcommandx{\telmo}[2][1=]{\vspace{0.2cm} \todo[linecolor=gold,backgroundcolor=gold!25,bordercolor=gold, #1]{\textbf{Telmo:} #2}}
\newcommandx{\antonio}[2][1=]{\vspace{0.2cm} \todo[linecolor=darkgreen,backgroundcolor=darkgreen!25,bordercolor=darkgreen, #1]{\textbf{Antonio:} #2}}
\newcommandx{\yannick}[2][1=]{\vspace{0.2cm} \todo[linecolor=yellow,backgroundcolor=yellow!25,bordercolor=yellow, #1]{\textbf{Yannick:} #2}}

\newcommand{\todotemplate}{\textcolor{red}{TODO}\xspace}
\newcommandx{\thendra}[2][1=]{\vspace{0.2cm} \todo[linecolor=blue,backgroundcolor=blue!15,bordercolor=blue, #1]{\textbf{\todotemplate @Hendra:} #2}}
\newcommandx{\ttelmo}[2][1=]{\vspace{0.2cm} \todo[linecolor=gold,backgroundcolor=gold!25,bordercolor=gold, #1]{\textbf{\todotemplate @Telmo:} #2}}
\newcommandx{\tantonio}[2][1=]{\vspace{0.2cm} \todo[linecolor=darkgreen,backgroundcolor=darkgreen!25,bordercolor=darkgreen, #1]{\textbf{\todotemplate @Antonio:} #2}}
\newcommandx{\tyannick}[2][1=]{\vspace{0.2cm} \todo[linecolor=yellow,backgroundcolor=yellow!25,bordercolor=yellow, #1]{\textbf{\todotemplate @Yannick:} #2}}

\makeatletter
\def\adl@drawiv#1#2#3{%
        \hskip.5\tabcolsep
        \xleaders#3{#2.5\@tempdimb #1{1}#2.5\@tempdimb}%
                #2\z@ plus1fil minus1fil\relax
        \hskip.5\tabcolsep}
\newcommand{\cdashlinelr}[1]{%
  \noalign{\vskip\aboverulesep
           \global\let\@dashdrawstore\adl@draw
           \global\let\adl@draw\adl@drawiv}
  \cdashline{#1}
  \noalign{\global\let\adl@draw\@dashdrawstore
           \vskip\belowrulesep}}
\makeatother

\makeatletter
\newcommand{\ssymbol}[1]{^{\@fnsymbol{#1}}}

\makeatother

\title{One Wide Feedforward is All You Need}

\author{
    \hspace{-35mm}Telmo Pessoa Pires$^{*\ssymbol{2}}$ \\
    \hspace{-35mm}Equall \\
    \hspace{-35mm}\texttt{telmo@equall.ai} \\
    \And
    \hspace{-20mm}António V. Lopes \quad Yannick Assogba \quad Hendra Setiawan$^*$ \\
    \hspace{-20mm}Apple \\
   \hspace{-20mm}\texttt{\{antoniovilarinholopes, yassogba, hendra\}@apple.com}
}

\begin{document}
\maketitle
\def\thefootnote{*}\footnotetext{Equal contribution.}\def\thefootnote{\arabic{footnote}}
\def\thefootnote{$\ssymbol{2}$}\footnotetext{Work conducted while at Apple.}\def\thefootnote{\arabic{footnote}}
\begin{abstract}
The Transformer architecture has two main non-embedding components: Attention and the Feed Forward Network (FFN). Attention captures interdependencies between words regardless of their position, while the FFN non-linearly transforms each input token independently. In this work we explore the role of the FFN, and find that despite taking up a significant fraction of the model's parameters, it is highly redundant. Concretely, we are able to substantially reduce the number of parameters with only a modest drop in accuracy by removing the FFN on the decoder layers and sharing a single FFN across the encoder. Finally we scale this architecture back to its original size by increasing the hidden dimension of the shared FFN, achieving substantial gains in both accuracy and latency with respect to the original Transformer Big.
\end{abstract}

\section{Introduction}

The Transformer architecture \citep{NIPS2017_attention} has become the de facto paradigm in many Natural Language Processing (NLP) tasks, including Machine Translation (MT).
Several studies have shown that Transformers exhibit impressive scaling-law properties \cite{gordon-etal-2021-data,pmlr-v162-bansal22b,ghorbani2022scaling}, wherein increasing the number of model parameters leads to further accuracy gains.
In parallel with this architecture's impressive scaling of the numbers of parameters \citep{chowdhery2022palm}, there is a growing trend towards reducing model footprints for real-world deployment, to satisfy practical constraints like latency requirements as well as memory and disk space limitations. 
In turn, researchers are actively exploring parameter sharing \citep{ge-etal-2022-edgeformer,takase-kiyono-2023-lessons,lou2022dictformer}, reducing the dimensionality of Transformer components, and pruning components like attention heads \citep{voita-etal-2019-analyzing,NEURIPS2019_2c601ad9}.

Although the role of attention in learning pairwise dependencies between tokens is relatively well understood \cite{voita-etal-2019-analyzing,clark-etal-2019-bert,vig-belinkov-2019-analyzing}, the role of the Feed Forward Network (FFN) remains under-explored.
Recently, \citet{geva-etal-2021-transformer} established a connection between the FFN and attention by positing that the FFN corresponds to learnable \emph{key}-\emph{value} pairs where the weights of the first layer of the FFN corresponds to the \emph{keys} and those of the second to the \emph{values}. 
They find that the keys are able to capture salient textual patterns at each layer, and they notice that the classes of patterns tend to overlap between neighboring layers, indicating redundancy in the representation. 

This observation motivates our work, where we revisit the conventional practice of allocating an individual FFN per layer. 
We investigate the effect of sharing and dropping the FFN across different layers on MT models.
We conduct thorough experiments with different configurations of the Transformer, across different language pairs, including a low resource language pair and multilingual. 
In addition, we investigate the effect of the FFN in a decoder-only Transformer-based model.
We find that a considerable level of redundancy exists between the encoder and decoder FFNs.
As a result, we are able to eliminate the decoder FFN and share a single FFN across the encoder without significantly compromising the model's accuracy. 
This step leads not only to significant parameter savings but also opens up opportunities for further improvements.
We also suggest using wider FFNs in the encoder while dropping the decoder's FFN, which results in a model with a similar size, but improved accuracy and reduced latency. 

Finally we conduct a fine-grained analysis of the representational similarity between the original model, using one independent FFN per layer, and various models with shared FFNs.
Our results reveal that both model accuracy and the internal representation of Transformer blocks remain stable when sharing the FFN.

\section{Background and Methodology}
\subsection{Transformer}
The Transformer architecture has two main components: attention and the FFN, which are connected via a residual connection \citep{kaiming_residualconnection} and layer normalization \citep{ba2016layer}.
In an encoder-decoder model, there are two types of attention: self-attention and cross-attention. 
Self-attention is used in both the encoder and the decoder, allowing the model to focus on relevant information within the same sequence. 
Cross-attention is exclusive to the decoder and allows it to attend to the encoder's output.
Attention takes as input a set of \emph{queries}, \emph{keys} and \emph{values}, projected using four $\mathbb{R}^{d_\text{model}\times d_\text{model}}$ matrices (one for the queries, keys, values, and final output) where $d_\text{model}$ is the model's hidden dimension. It then applies the \textsc{Softmax} function to allow it to focus on the most relevant values.

The FFN is applied after attention on both the encoder and the decoder and consists of the following $2$-layer linear transformation:
\begin{equation}
    \text{FFN}(\boldsymbol{x}) = \max(0, \boldsymbol{x}\mathbf{W}_1 + b_1)\mathbf{W}_2 + b_2,
    \label{eq:ffn}
\end{equation}
where a \textsc{Relu} non-linearity is applied to the transformation of the input sequence ($\boldsymbol{x}$). 
At each layer, the FFN is parameterized with two matrices, $\mathbf{W}_1 \in \mathbb{R}^{d_\text{model}\times d_\text{ff}}$ and $\mathbf{W}_2 \in \mathbb{R}^{d_\text{ff} \times d_\text{model}}$ where $d_\text{ff}$ is the \emph{FFN dimension} and is usually set to $4\times d_\text{model}$ \citep{NIPS2017_attention}. 

Recent work has drawn a significant link between attention and the FFN \citep{geva-etal-2021-transformer}, wherein $\mathbf{W}_1$ and $\mathbf{W}_2$ assume roles akin to the \emph{keys} and \emph{values} to an unnormalized attention where the input ($\boldsymbol{x}$) acts as the \emph{query}. 
Unlike regular attention, the FFN employs a \textsc{Relu}, which allows \emph{multiple} keys to significantly contribute to the final output \citep{geva-etal-2021-transformer}.
Additionally, these keys correspond to an inventory of salient patterns that are learned from the training data.
\citet{geva-etal-2021-transformer} suggest that at the lower layers the FFN learns shallow syntactic patterns and progressively learns deep semantic patterns on the deeper layers. Moreover, the authors find that there's a substantial overlap between patterns captured by adjacent layers, indicating that there are redundancies in the FFNs and suggesting a better allocation of these parameters might be beneficial for performance.

\subsection{Sharing and Widening the FFN}
The vanilla Transformer allocates one FFN for each layer of the encoder and decoder, \ie $\text{FFN}^{enc}_{i}$ or $\text{FFN}^{dec}_{i}$, respectively. 
Excluding embedding parameters, these FFNs occupy around two thirds of the parameter budget, while attention occupies the remaining third\footnote{Ignoring layer normalization, there are $4\times d_\text{model}\times d_\text{model}$ parameters for attention vs $2\times d_\text{model}\times d_\text{ff} = 8\times d_\text{model}\times d_\text{model}$ parameters for the FFN, assuming $d_\text{ff} = 4\times d_\text{model}$.}.
Earlier work found that constraining the parameterization of the decoder FFNs causes no degradation in accuracy \cite{ge-etal-2022-edgeformer}.
In this work, we share the parameters of the FFN across layers and/or across the encoder and decoder to minimize redundancy between FFNs.

Let $N_{enc}, N_{dec}$ be the numbers of encoder and decoder layers, respectively. We consider multiple configurations for parameter sharing as follows:
\begin{itemize}
    \item One $\textbf{FFN}^{enc}_{all}$ for the whole encoder: 
    \item[] $\text{FFN}^{enc}_{i}(\cdot) \overset{\text{tied}}{=} \text{FFN}^{enc}_{all}(\cdot), \forall i : 1 \leq i \leq N_{enc}$
    \item One $\textbf{FFN}^{dec}_{all}$ for the whole decoder: 
    \item[] $\text{FFN}^{dec}_{j}(\cdot) \overset{\text{tied}}{=} \text{FFN}^{dec}_{all}(\cdot), \forall j : 1 \leq j \leq N_{dec}$
    \item One $\textbf{FFN}^{encdec}_{all}$ for both the encoder and the decoder: 
    \item[] $\text{FFN}^{enc}_{i}(\cdot) \overset{\text{tied}}{=} \text{FFN}^{dec}_{j}(\cdot) \overset{\text{tied}}{=} \text{FFN}^{encdec}_{all}(\cdot),$
    \item[] \quad\quad $\forall i, j : 1 \leq i \leq N_{enc}, 1 \leq j \leq N_{dec}$
\end{itemize}

Additionally, we explore modifying the dimension of the shared FFN, which we denote as $d_{\text{ff}^\prime}$. 
Setting $d_{\text{ff}^\prime} > d_\text{ff}$ widens the shared FFN while $d_{\text{ff}^\prime} < d_\text{ff}$ narrows it.  
We also consider the extreme cases of setting $d_{\text{ff}^\prime}$ to $0$ or to $(N_{enc} + N_{dec}) \times d_\text{ff}$ (and beyond). 
Setting $d_{\text{ff}^\prime}=0$ is equivalent to dropping the FFN\footnote{In our experiments without the FFN (\ie, $d_{\text{ff}^\prime} = 0$) we remove the residual connection and layer normalization associated with it, as they become redundant.} while setting $d_{\text{ff}^\prime}=(N_{enc} + N_{dec}) \times d_\text{ff}$ is akin to sharing the concatenation of all individual FFNs.

Sharing the FFNs directly affects the number of parameters and, to a certain extent, latency. 
For instance, sharing $\text{FFN}^{enc}_{all}$ for the whole encoder reduces the number of parameters by $(N_{enc} - 1) \times 2 \times d_\text{model} \times d_\text{ff}^\prime$\footnote{Plus the layer normalization parameters, which we are ignoring for simplicity.}; whereas 
removing the FFN on the decoder, \ie, setting $d_{\text{ff}^\prime} = 0$ for $\text{FFN}^{dec}_{all}$, reduces the parameters by $(N_{dec}) \times 2 \times d_\text{model} \times d_\text{ff}^\prime$ and reduces the amount of computation to be done.
This is particularly important during inference since the forward pass of the decoder is autoregressive, and changing the decoder's FFN dimension has a higher latency impact than on the encoder.

Since different configurations have different impacts, we analyse the trade-off between model size, latency, and accuracy:
\begin{enumerate*}[label=(\roman*)]
    \item How many parameters can be shared/pruned with negligible (if any) accuracy degradation?
    \item Are the encoder and decoder FFNs affected similarly?
    \item Keeping the same model size, can the FFN parameters be allocated more efficiently?
\end{enumerate*}

We propose a novel configuration, which we call the \emph{One Wide FFN} model, consisting of a single shared wide FFN on the encoder and no FFN on the decoder.
To keep the number of parameters the same as in the baseline, we increase the shared FFN dimension accordingly: $\text{FFN}^{enc}_{all}$ with $d_{\text{ff}^\prime}=(N_{enc} + N_{dec}) \times d_\text{ff}$. 

For completeness, we include similar experiments on the attention mechanism in \Cref{app:attention}. These experiments show that, contrary to the FFN, individual layer-specific attention weights are more important and not as redundant, as sharing the attention leads to significant accuracy drops.

\subsection{Representational Similarity}
\label{sec:visualizations}

Besides investigating the impact on accuracy, we study the similarity between different models in terms of their \emph{internal representations} and the \emph{semantic space} they produce.

We use Linear Centered Kernel Alignment (\cka, \citealp{pmlr-v97-kornblith19a}) to measure the similarity between the internal representations of different models. 
\cka uses inner products to estimate how similar the kernel matrices of two different representations are, and is based on the Hilbert-Schmidt Independence Criterion (HSIC, \citealp{gretton2005measuring}), a statistical measure of independence of two random variables. 
Linear \cka uses the dot product as a kernel and can be written as:
\begin{align}
    \textsc{CKA}(\mathbf{A}, \mathbf{B}) &= \frac{||\mathbf{A}\mathbf{B}^\text{T}||_\text{F}^2}{||\mathbf{A}^\text{T}\mathbf{A}||_\text{F} ||\mathbf{B}^\text{T}\mathbf{B}||_\text{F}}, \nonumber
\end{align}
where $||\cdot||_\text{F}$ is the Frobenius norm while $\mathbf{A}$ and $\mathbf{B}$ are mean-centered (\ie, we subtract the mean) feature matrices of the layers under comparison, computed on the same dataset. Both matrices are $n\times d$, where $n$ is the number of sentences in the dataset and $d$ is the output dimension of the component, and are obtained by averaging the activation of all tokens in each sentence\footnote{We use the source sentence and force decode the first reference to compute the encoder and decoder representations, respectively.}. The linear kernel is straightforward to compute and \citealp{pmlr-v97-kornblith19a} report strong empirical performance of linear \cka compared to other kernels and methods.

To measure the similarity between the semantic spaces of different models, we use Local Neighborhood Similarity (\lns, \citealp{boggust_embedding_2022}).
Local neighborhood similarities have been previously been used in analyzing semantic shifts in word embeddings \citep{hamilton-etal-2016-cultural}. 
The premise of \lns is that two semantic spaces are similar if a sentence has similar neighbors in the two spaces.
The \lns of a sentence $s$ between models $1$ and $2$ is defined as:
\begin{equation}
    \textsc{LNS}(s) = Sim(\knn_1(s),\;\knn_2(s)),
    \label{eq:similarity} \nonumber
\end{equation}
where $\knn(s)$ is the set of $k$ nearest neighbors of sentence $s$ for a model and $Sim$ is the intersection-over-union (Jaccard similarity) of the two sets of neighbors. 
For each pair of components (attention and FFN) in models $1$ and $2$ we compute the \lns of all sentences in the evaluation dataset and take the mean \lns as our layer similarity measure. 
The smaller the value of $k$ the more local the neighborhoods we are comparing, and the more specific the retrieval task. 
We pick $k$ to be small enough to visually inspect sentence neighborhoods if necessary.
In our analysis, we use cosine distance as the distance metric between activations and set $k$ to $5\%$ of the dataset size ($\sim 100$ sentences).

\section{Experimental Setup}
\label{sec:experimental_setup}

\paragraph{Data}
In our experiments, we show results on WMT22 English (\len) $\rightarrow$ German (\lde) ($296$M pairs), which we obtained using the provided \texttt{mt-data} scripts\footnote{\url{https://www.statmt.org/wmt22/mtdata/}}, WMT16
\len$\rightarrow$ Romanian (\lro) ($610$K pairs), and for the multilingual setup of \citet{pires-etal-2023-learning}, consisting of $10$ languages: German, English, Spanish, French, Italian, Japanese, Korean, Portuguese, Swahili, and Chinese. 
In our analysis, we mostly focus on WMT22 \len$\rightarrow$\lde.

Following \citet{schmidt-etal-2022-non}, we use WMT'16 provided scripts to normalize the \lro{} side.
\len$\rightarrow$\lro keeps diacritics for producing accurate translations. 
For more details refer to \citet{schmidt-etal-2022-non}. For the multilingual experiments, we replicated the setup of \citet{pires-etal-2023-learning}, which includes all details, including data preprocessing and dataset sizes.

\paragraph{Metrics}
We compute \bleu\footnote{For the multilingual experiments, we select the Flores101 tokenizer in \sacrebleu, so technically we report \spBLEU.} using \sacrebleu\footnote{\url{https://github.com/mjpost/sacrebleu}} version $2.3.1$, with evaluation signatures \texttt{nrefs:1} \texttt{|} \texttt{case:mixed} \texttt{|} \texttt{eff:no} \texttt{|} \texttt{tok:13a} \texttt{|} \texttt{smooth:exp} for \bleu, and \texttt{nrefs:1} \texttt{|} \texttt{case:mixed} \texttt{|} \texttt{eff:no} \texttt{|} \texttt{tok:flores101} \texttt{|} \texttt{smooth:exp} for \spBLEU. For our main results, we also report \comet using the \texttt{wmt20-comet-da} model and \chrf using the signature \texttt{nrefs:1} \texttt{|} \texttt{case:mixed} \texttt{|} \texttt{eff:yes} \texttt{|} \texttt{nc:6} \texttt{|} \texttt{nw:0} \texttt{|} \texttt{space:no}.

\paragraph{Latency}
We report inference time in tokens/second (the higher, the better), averaged over $5$ runs. For the multilingual models, we use the \lde$\rightarrow$\len test set. Our measurements were collected using a single NVIDIA V100 GPU on a single-threaded Intel(R) Xeon(R) Gold 6148 CPU @ 2.40GHz with batch size of 1 and beam size of 5, in order to realistically mimic the inference of a deployed model. 
For experiments with larger batch sizes, see \Cref{app:batch_size}.

\paragraph{Tokenization}
For WMT22 \len$\rightarrow$\lde, we use \textsc{SentencePiece} \citep{kudo-richardson-2018-sentencepiece}, with a vocabulary size of $32$K and a character coverage of $1.0$, while for the multilingual experiments we use a vocabulary size of $250$k and a character coverage of $0.9995$. 
For WMT16 \len$\rightarrow$\lro we use byte-pair encoding (BPE, \citealp{sennrich-etal-2016-neural}) with $40,000$ merge operations.

\paragraph{Model Architectures}
We focus our analysis on the Transformer \texttt{Big}  where $N_{enc} = N_{dec} = 6$, $d_{model} = 1024$, $d_\text{ff} = 4096$, and it has $16$ attention heads. 
We also report results on Transformer \texttt{Base} ($N_{enc} = N_{dec} = 6$, $d_{model} = 512$, $d_\text{ff} = 2048$, and $8$ attention heads), and a deep encoder shallow decoder \citep{Kasai0PCS21} Transformer Big with $12$ encoder layers, and $2$ decoder layers. 
For our decoder-only experiments, the model is identical to the Transformer Big, except that all $12$ layers are on the decoder. 
Our decoder-only model is similar to a Transformer-based language model, particularly Prefix-LM \cite{JMLR:v21:20-074}, where we apply a non-autoregressive mask on the source side and an autoregressive mask on the target.
The source and target embeddings and the output projection matrix are shared in all models \cite{press-wolf-2017-using}.

\paragraph{Hyperparameters} All experiments are implemented using \textsc{fairseq} \citep{ott-etal-2019-fairseq}. Our optimizer is \adam \citep{KingmaB14} with a learning rate of $0.0007$.
We train for $80$k, $80$k, $150$k steps on WMT22, WMT16, and multilingual, respectively, at which point the models had converged. We use $4000$ warm-up steps, and an inverse square root learning rate scheduler \citep{NIPS2017_attention}. We use a dropout rate of $0.1$ for WMT22, $0.3$ for WMT16, and $0$ for the multilingual experiments due to the abundance of data, following \citet{pires-etal-2023-learning}. All models are trained using \texttt{fp16} \citep{ott-etal-2018-scaling}.

\paragraph{Nomenclature}
In our experiments, we run a number of different configurations per model architecture that differ in the way the FFN is used, shared, or dropped, as well the size of the shared FFN ($d_{\text{ff}^\prime}$). 
To facilitate our discussion, we introduce in \Cref{tab:nomenclature} the nomenclature that will serve as reference for the rest of the text. 
Unless otherwise stated, the dimension of the shared $\text{FNN}_{all}^{\ast}$, \ie $d_{\text{ff}^\prime}$ is equal to the $d_\text{ff}$ of the original model.

For decoder-only models, only \texttt{SharedDec} and \texttt{NoDec} configurations are defined.
For conciseness, we drop the mention of FFN from the text when possible, i.e. \texttt{SharedEnc} instead of \texttt{SharedEncFFN}.

\begin{table}[ht!]
    \centering
    {\setlength{\tabcolsep}{1.5pt}
    \begin{tabular}{lcc}
    \toprule              
    FFN Description & Encoder & Decoder \\
    \midrule
    \texttt{SharedEnc} & $\text{FNN}_{all}^{enc}$ & $\text{FNN}_{i}^{dec}$ \\
    \texttt{SharedDec} & 
    $\text{FNN}_{i}^{enc}$ & $\text{FNN}_{all}^{dec}$ \\    
    \texttt{SharedEncSharedDec} & 
    $\text{FNN}_{all}^{enc}$ & $\text{FNN}_{all}^{dec}$ \\    
    \texttt{SharedEncDec} & 
    \multicolumn{2}{c}{$\text{FNN}_{all}^{encdec}$} \\    
    \texttt{NoDec} & 
    $\text{FNN}_{i}^{enc}$ & \texttt{No-op} \\   \texttt{SharedEncNoDec} & 
    $\text{FNN}_{all}^{enc}$ & \texttt{No-op} \\   
    \bottomrule
    \end{tabular}}
    \caption{Nomenclature used in our experiments. \texttt{No-op} indicates an identity function, which is equivalent to dropping the FFN.}
    \label{tab:nomenclature}
\end{table}

\paragraph{Representational Similarity} We use the WMT22 \len$\rightarrow$\lde evaluation set for both \cka and \lns analysis. 
We analyze encoder and decoder representations independently and present these metrics in a matrix heatmap plot showing pairwise similarity between layers. 
The diagonal of this matrix is the similarity of corresponding layers, \ie, layer $i$ on both architectures.
In order to facilitate an ``apples-to-apples'' comparison across models, we extract decoder representations by force decoding the (first) reference.
We establish $2$ crucial similarity scores: 
a \emph{benchmark} on similarity for each of these metrics, where we train two additional models using the same architecture but with different random seeds;
a similarity lower bound, where we compare the baseline Transformer Big with a randomly initialized (\ie, untrained) model with the same architecture. 
We present these bounds in \Cref{app:appendix_similarity}.

\section{Experimental Results}
\subsection{Sharing FFNs}
\label{sec:ffn_sharing}

The results of various FFN sharing configurations are summarized in \Cref{tab:sharing-enc-dec-ffn}, including their impact on accuracy and model size (in millions of parameters and percentage). 
Sharing either the encoder (\texttt{SharedEnc}) or the decoder FFN (\texttt{SharedDec}) results in just a $0.2$ to $0.3$ \bleu point decrease, while reducing the parameter count by nearly $20\%$.
Sharing the FFN on each side (\texttt{ShareEncShareDec}) leads to a more substantial degradation of $0.9$ \bleu points, albeit reducing the parameter count by $37\%$, while sharing a single FFN on the encoder and decoder (\texttt{ShareEncDec}) results in a slightly higher degradation of $1.1$ \bleu{} points.
Nonetheless, these findings support the hypothesis that the FFN contains some degree of redundancy, as we expected a greater accuracy degradation given the substantial ($20-40\%$) reduction in model size.

\begin{table}[ht!]
    \centering
    {\setlength{\tabcolsep}{1.2pt}
    \begin{tabular}{lccr@{}r}
    \toprule              
    Architecture & \bleu & \multicolumn{2}{c}{$\mid\theta\mid~(\%)$} \\
    \midrule
    \texttt{Transformer Big} & $35.6$ & $228$M & ($100$) \\
    \quad + \texttt{SharedEnc} & $35.4$ & $186$M & ($82$) \\
    \quad + \texttt{SharedDec} & $35.3$ & $186$M & ($82$) \\
    \quad + \texttt{SharedEncSharedDec} & $34.7$ & $144$M& $(63)$ \\
    \quad + \texttt{SharedEncDec} & $34.5$ & $136$M& ($59$) \\
    \bottomrule
    \end{tabular}}
    \caption{\sacrebleu results on WMT 22 \len$\rightarrow$\lde for different FFN sharing configurations. $\mid\theta\mid$ is the number of parameters.}
    \label{tab:sharing-enc-dec-ffn}
\end{table}

While we focus on sharing \emph{one} FFN for all layers within a module, we compare with sharing multiple FFNs following \citet{takase-kiyono-2023-lessons} in \Cref{app:custom_sharing}. We find that sharing one FFN is as accurate as sharing multiple FFNs within a module, while being more parameter-efficient.

\subsection{Dropping FFNs}
\label{sec:ffn_dropping}

\Cref{tab:no-enc-dec-ffn} summarizes the performance of models with no FFNs. Besides \bleu and number of parameters, we report the inference speed for each architecture.
Dropping the FFN on the encoder (\texttt{NoEnc}) leads to a $0.9$ \bleu point drop while reducing the parameter count by $22\%$ and with minimal effect on inference speed.
Dropping the FFN on the decoder (\texttt{NoDec}), on the other hand, causes a degradation of only $0.4$ \bleu points while increasing the inference speed by $20\%$\footnote{The reason for this difference between \texttt{NoEnc} and \texttt{NoDec} is that the encoder output is computed in parallel, while the decoder operates in a step-by-step fashion.}.
The highest latency reduction is obtained by removing the FFNs on both the encoder and the decoder (\texttt{NoEncNoDec}), but it comes with a significantly larger degradation of over $2$ \bleu points.

\begin{table}[ht!]
    \centering
    {\setlength{\tabcolsep}{1.2pt}
    \begin{tabular}{lcccr@{}r}
    \toprule              
    Architecture & \bleu & Speed & \multicolumn{2}{c}{$\mid \theta\mid~(\%)$} \\
    \midrule
    \texttt{Transformer Big} & $35.6$ & $111^{\pm 1.2}$ & $228$M & $(100)$ \\
    + \texttt{NoEnc} & $34.7$ & $112^{\pm 1.0}$ & $178$M & $(78)$ \\
    + \texttt{NoDec} & $35.2$ & $133^{\pm 0.9}$ & $178$M & $(78)$ \\
     + \texttt{NoEncNoDec} & $33.5$ &  $138^{\pm 1.9}$  & $127$M & $(56)$ \\
    \cdashlinelr{1-4}
     + \texttt{SharedEncNoDec} & $35.3$ &  $136^{\pm 1.1}$   & $136$M & $(60)$ \\
     + \texttt{NoEncSharedDec} & $33.9$ &   $127^{\pm 1.0}$   & $136$M & $(60)$ \\
    \bottomrule
    \end{tabular}}
    \caption{\sacrebleu results on WMT 22 \len$\rightarrow$\lde for different FFN dropping configurations.}
    \label{tab:no-enc-dec-ffn}
\end{table}

\paragraph{Combining sharing and dropping}
These results, together with those from \Cref{tab:sharing-enc-dec-ffn}, suggest that the encoder and decoder FFNs have different contributions: the decoder's are more redundant, corroborating previous work on FFNs parametrization \cite{ge-etal-2022-edgeformer}.
With this in mind, we experiment with one shared FFN on the encoder and dropping it on the decoder, reported as \texttt{SharedEncNoDec} in \Cref{tab:no-enc-dec-ffn}.
As shown, with just approximately $60\%$ of Transformer Big parameters we observe a $22\%$ improvement in inference speed, at the cost of $0.3$ \bleu point.

\begin{table*}[ht]
    \centering
    \begin{tabular}{l@{\quad}r@{.}lr@{.}lr@{.}lr@{.}lr@{\,\,}r}
    \toprule              
     & \multicolumn{2}{c}{\bleu} & \multicolumn{2}{c}{\chrf} & \multicolumn{2}{c}{\comet} & \multicolumn{2}{c}{Speed} & \multicolumn{2}{c}{$\mid \theta\mid~(\%)$} \\    \midrule
    Transformer Big \len$\rightarrow$\lde  &   $35$&$6$ & $62$&$6$ & $57$&$2$ & $110$&$8^{\pm 1.2}$ & $228$M & $(100)$ \\
    \quad + \texttt{SharedEncNoDec} FFN $d_{\text{ff}^\prime}=4,096$ & $35$&$3$ & $62$&$1$ & $56$&$1$ & $ 135$&$7^{\pm 1.1}$ & $135$M & $(60)$  \\
    \quad + \texttt{SharedEncNoDec} FFN $d_{\text{ff}^\prime}=24,576$ & $35$&$7$ & $62$&$7$ & $57$&$9$ & $ 138$&$2^{\pm 0.9}$ & $177$M & $(80)$  \\
    \quad + \texttt{SharedEncNoDec} FFN $d_{\text{ff}^\prime}=49,152$ & $\textbf{36}$&$\textbf{5}^\dag$ & $\textbf{63}$&$\textbf{2}^\dag$ & $\textbf{59}$&$\textbf{6}$ & $137$&$5^{\pm 1.6}$ & $228$M & $(100)$  \\
    \quad + \texttt{SharedEncNoDec} FFN $d_{\text{ff}^\prime}=98,304$ & $36$&$4^\dag$ & $\textbf{63}$&$\textbf{2}^\dag$ & $59$&$0$ & $134$&$5^{\pm 1.6}$ & $328$M & $(145)$  \\
    \bottomrule
    \end{tabular}
    \caption{Accuracy of One Wide FFN for Transformer Big \len$\rightarrow$\lde on WMT$22$. $\dag$ implies the system is statistical significantly different at $p < 0.05$.}
    \label{tab:full-capacity-results:a}
\end{table*}

\subsection{One Wide FFN Model}
\label{sec:one-wide-ffn}
Previous sections describe models that share and/or drop FFNs, effectively reducing model size at some modest accuracy cost. 
In this section, we investigate whether we can regain the accuracy lost while preserving the parameter efficiency and the latency reduction.
We focus on \texttt{ShareEncNoDec} model as it provides a strong baseline with significant parameter savings and inference speedups.

We propose increasing the dimension of the shared FFN to match the number of parameters of the original (fully-parameterized) model, so as to avoid increasing the overhead of model storage.
In particular, \texttt{ShareEncNoDec} saves around $(N_{enc}+N_{dec}-1) \times 2 \times d_{model} \times d_\text{ff}$ parameters as there's one single shared FFN in the encoder.
On the other hand, the \texttt{Transformer Big} has $(N_{enc}+N_{dec})$ FFNs.
Thus, we match the size of the original model by setting the dimension of the shared FFN, $d_{\text{ff}^\prime}$, to $(N_{enc}+N_{dec}) \times d_\text{ff}$.

\Cref{{tab:full-capacity-results:a}} summarizes our results. It includes our proposed model, the \emph{One Wide FFN} model ($d_{\text{ff}^\prime} = 49,152$), as well as the baseline \texttt{Transformer Big}, and the corresponding \texttt{ShareEncNoDec} ($d_{\text{ff}^\prime} = 4,096$). It also includes a wide model with $d_{\text{ff}^\prime} = 24,576$, which uses the same number of parameters as \texttt{NoDec}, with $d_{\text{ff}^\prime} = N_{enc} \times d_\text{ff}$. This model achieves an accuracy on par (or slightly above) the baseline \texttt{Transformer Big} with $20\%$ fewer parameters and a significant inference speed-up. 

Our proposed model with $d_{\text{ff}^\prime}=49,152$ goes beyond that, achieving a gain of $1.2$ \bleu points over the vanilla \texttt{ShareEncNoDec} and $0.9$ \bleu points over the \texttt{Transformer Big}. 
These gains remain consistent across \chrf and \comet. 
Furthermore, it has a similar inference speed as the \texttt{ShareEncNoDec} model. 
For completeness, we include a wider model with $d_{\text{ff}^\prime}=98,304$. 
Despite the extra capacity, this model does not provide any additional accuracy gains, which we suspect is due to the lack of data to train such a large model.

\subsection{Analyzing Internal Representations}
\label{sec:internal_representation}

We now report a \emph{post-hoc} analysis of the internal representations of the models introduced in preceding sections. 
Our objectives are twofold: 1) to ascertain whether the proposed models' internal representations exhibit a significant degree of similarity to those of the original base model; 2) to delve into the impact of the proposed methods on redundancy.
We adopt the definition of redundancy of \newcite{dalvi-etal-2020-analyzing}, who \emph{visually} inspect the similarity between adjacent modules within a model (high similarity entails high redundancy).

\begin{table}[ht!]
    \centering
    {\setlength{\tabcolsep}{1.2pt}
    \begin{tabular}{lr@{.}lr@{.}lr@{.}lr@{.}l}
    \toprule
    \multirow{2}*{Architecture} & \multicolumn{4}{c}{Encoder} & \multicolumn{4}{c}{Decoder} \\
    & \multicolumn{2}{c}{\cka} & \multicolumn{2}{c}{\lns} & \multicolumn{2}{c}{\cka}  & \multicolumn{2}{c}{\lns} \\
    \midrule
    \texttt{Benchmark} & 100 & 0 & 100 & 0 & 100 & 0 & 100 & 0 \\
    \midrule
    \texttt{SharedEnc} & \cellcolor{gray!25}98 & \cellcolor{gray!25}$0$ & \cellcolor{gray!25}$96$ & \cellcolor{gray!25}$2$ & $100$ & $8$ & $100$ & $6$ \\
    \texttt{SharedDec}  & $100$ & $2$ & $101$ & $4$ & \cellcolor{gray!25}$98$ & \cellcolor{gray!25}$3$ & \cellcolor{gray!25}$94$ & \cellcolor{gray!25}$6$  \\
    \texttt{SharedEncSharedDec}  & \cellcolor{gray!25}$98$ & \cellcolor{gray!25}$9$ & \cellcolor{gray!25}$97$ & \cellcolor{gray!25}$2$ & \cellcolor{gray!25}$99$ & \cellcolor{gray!25}$5$ & \cellcolor{gray!25}$95$ & \cellcolor{gray!25}$4$ \\
    \texttt{SharedEncDec}  & \cellcolor{gray!25}$97$ & \cellcolor{gray!25}$6$ & \cellcolor{gray!25}$94$ & \cellcolor{gray!25}$4$ & \cellcolor{gray!25}$98$ & \cellcolor{gray!25}$4$ & \cellcolor{gray!25}$93$ & \cellcolor{gray!25}$5$ \\
    \texttt{NoEnc}  & \cellcolor{blue!25}$90$ & \cellcolor{blue!25}$0$ & \cellcolor{blue!25}$70$ & \cellcolor{blue!25}$5$ & $101$ & $0$ & $96$ & $8$  \\    
    \texttt{NoDec}  & $100$ & $0$ & $98$ & $6$ & \cellcolor{blue!25}$96$ & \cellcolor{blue!25}$0$ & \cellcolor{blue!25}$87$ & \cellcolor{blue!25}$4$  \\    
    \texttt{SharedEncNoDec} & \cellcolor{gray!25}$97$ & \cellcolor{gray!25}$6$ & \cellcolor{gray!25}$98$ & \cellcolor{gray!25}$9$ & \cellcolor{blue!25}$97$ & \cellcolor{blue!25}$5$ & \cellcolor{blue!25}$89$ & \cellcolor{blue!25}$0$ \\
    \midrule
    \texttt{SharedEncNoDec}$^{d_\text{ff}^\prime=49152}$ & \cellcolor{gray!25}$97$ & \cellcolor{gray!25}$0$ & \cellcolor{gray!25}$83$ & \cellcolor{gray!25}$2$ & \cellcolor{blue!25}$94$ & \cellcolor{blue!25}$0$ & \cellcolor{blue!25}$82$ & \cellcolor{blue!25}$9$ \\    
    \bottomrule
    \end{tabular}}
    \caption{Similarity of the representations ($\%$) of corresponding modules of different architectures vs. the Transformer Big for WMT22 \len$\rightarrow$\lde. These scores are normalized by comparing them to the \cka and \lns benchmark scores. For \texttt{NoDec} configurations we compare the final output of the Transformer layer as a whole as they have different modules than the baseline. The columns for shared and for dropped FFNs are highlighted in \colorbox{gray!25}{gray} and \colorbox{blue!25}{blue} respectively.}
    \label{tab:sharing-enc-dec-ffn-similarity}
\end{table}

\subsubsection{Similarity to Baseline}

\begin{figure*}[ht]
    \centering
    \begin{subfigure}[t]{0.49\textwidth}
        \centering
        \includegraphics[width=\textwidth]{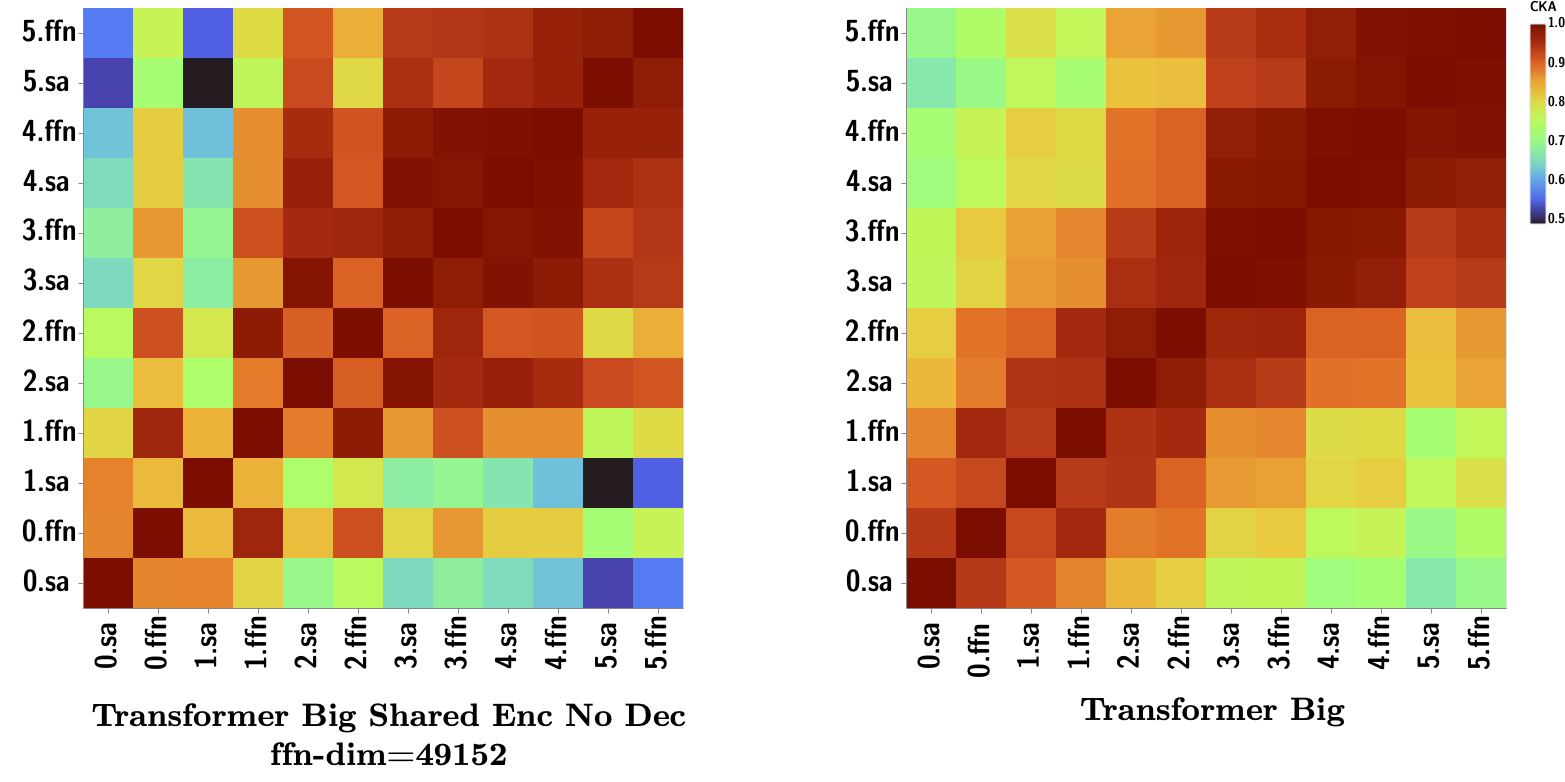}
        \caption{Encoder self similarity.}
        \label{fig:cka_one_wide_ffn:enc}
    \end{subfigure}
    \hfill
    \begin{subfigure}[t]{0.49\textwidth}
        \centering
        \includegraphics[width=\textwidth]{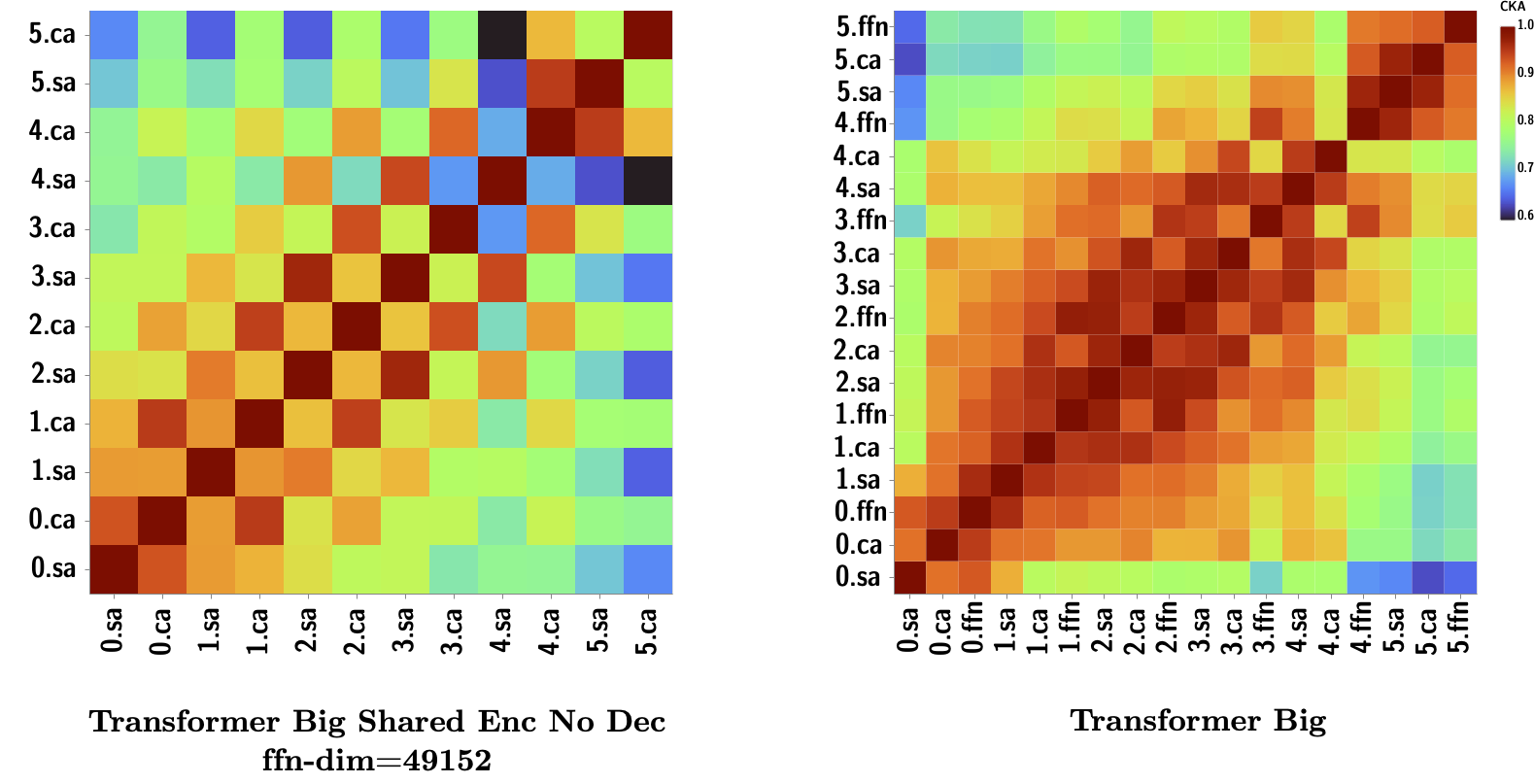}
        \caption{Decoder self similarity.}
        \label{fig:cka_one_wide_ffn:dec}
    \end{subfigure}
    \caption{\cka self similarity of encoder and decoder layers of the \textit{One Wide Encoder} model vs. the Transformer Big baseline. We identify each component with a label: \texttt{index.name}. For example, \texttt{0.sa} refers to the self-attention on layer $0$, while \texttt{4.ca} refers to the cross-attention on layer $4$.}
    \label{fig:cka_one_wide_ffn}
\end{figure*}

We ground the pairwise similarity metrics, by normalizing them against a benchmark.  
As mentioned in \Cref{sec:experimental_setup}, we establish the \emph{benchmark scores} by training two additional Transformer Big models, but using different random seeds. These models achieve similar accuracy as the baseline model (see \Cref{app:appendix_similarity:benchmark} for more details). The benchmark score is the similarity between the baseline and these models 
Because the benchmark is calculated by averaging similarity scores from different training runs of our baseline, individual runs can have a normalized score above $100\%$.

\Cref{tab:sharing-enc-dec-ffn-similarity} shows normalized similarity scores for several models.
Under the Encoder columns we compare the encoder representations, and under the Decoder columns we compare decoder representations.
Sharing FFNs leads to consistenly lower (normalized) similarity scores than models that do not share, both in terms of internal representation (\cka) and semantic spaces (\lns). 
As shown, although models that share FFNs have lower similarity scores compared to those that do not, the scores are still very close to $100\%$.
Moreover, these decreases align with the drops in \bleu seen in \Cref{tab:sharing-enc-dec-ffn}, where the model with the lowest similarity score (\texttt{ShareEncDec}) is also the least accurate model.
We observe a similar trend for models that drop the FFNs in the encoder or decoder, these models exhibit lower similarity scores with the respective component than models sharing them, as shown by \texttt{NoEnc} and \texttt{NoDec}.
In addition, the former result again suggests the FFNs in the encoder are more important than in the decoder as the similarity shifts drastically compared to all other settings.

For completeness, we report on the last row the similarity scores for the One Wide FFN model, which is more accurate than the base model.
The internal representations generated by that model diverge from those of the base model.
Interestingly, we observe a larger drop in \lns scores than in \cka scores, indicating that the shift occurs mostly in semantic space, rather than the Euclidean space captured by \cka.
For a detailed layer-wise similarity analysis that breaks out the aggregate analysis in \Cref{tab:sharing-enc-dec-ffn-similarity} see \Cref{app:appendix_similarity:layerwise}. 

\subsubsection{A Qualitative View of Redundancy}
We now study into the impact of our One Wide FFN model on the redundancy of the internal representations. 
In addition to adopting their definition of redundancy, we also adopt \citet{dalvi-etal-2020-analyzing}'s method of computing self-similarity, namely looking at how the representations change as they go through each module (self-attention, FFN, or cross-attention) of the model.
In particular, we use \cka to compute similarity between the output of different modules within the same model.

In \Cref{fig:cka_one_wide_ffn:enc}, we show the \cka self-similarity matrices for the encoders of the One Wide FFN model and the \texttt{Transformer Big}. We do the same for the decoders in \Cref{fig:cka_one_wide_ffn:dec}.
These matrices show how similar each module of the network is to all other modules \textit{within that network}. 
The diagonal of the matrix is the similarity between a module and itself and is always $1$.

As shown, there is high similarity between adjacent modules of the \texttt{Transformer Big}, both on the encoder and decoder, indicated by areas with darker red around the diagonal.
The prevalence of high similarity patterns among adjacent modules suggests a substantial degree of redundancy, and eliminating a module has a negligible impact on the final representations.
On the other hand, we observe a distinct checkerboard pattern on the self-similarity matrices of the One Wide FFN model, where individual modules tend to exhibit lower similarity with their immediate neighbors than with their second neighbors (\ie, the neighbors of the neighbors).
On the encoder, the checkerboard pattern emerges especially in the earlier modules while on the decoder, that pattern appears more consistently throughout the layers.
This pattern gives an indication that our model is learning non-trivial transformations of the input, leading to decreased redundancy within the network.

\subsection{Other architectures and Languages}

\begin{table*}[ht]
    \centering    
    \begin{tabular}{l@{\,\,}r@{.}lr@{.}lr@{.}lr@{.}lr@{\,\,}r}
    \toprule              
     & \multicolumn{2}{c}{\bleu} & \multicolumn{2}{c}{\chrf} & \multicolumn{2}{c}{\comet} & \multicolumn{2}{c}{Speed} & \multicolumn{2}{c}{$\mid \theta\mid~(\%)$} \\
    \midrule
    Transformer Base \len$\rightarrow$\lde  &  $34$ & $2$ & $61$ & $6$ & $54$ & $1$ & $116$ & $3^{\pm 0.9}$ & $70$M & $(100)$ \\
    \quad + \texttt{SharedEncNoDec} FFN $d_{\text{ff}^\prime}=2,048$ & $32$ & $5^\dag$ & $60$ & $1^\dag$ & $50$ & $0$ & $ 146$ & $0^{\pm 1.6} $ & $47$M & $(67)$  \\
    \quad + \texttt{SharedEncNoDec} FFN $d_{\text{ff}^\prime}=24,576$  & $\textbf{34}$ & $\textbf{7}$ & $\textbf{61}$ & $\textbf{8}$ & $\textbf{55}$ & $\textbf{6}$ & $ 146$ & $8^{\pm 1.3} $ & $70$M & $(100)$ \\
    \midrule
    Transformer Decoder-Only \len$\rightarrow$\lde &   $35$ & $8$ & $62$ & $8$ & $57$ & $7$ &  $ 79$ & $8^{\pm 1.9} $& $202$M & $(100)$ \\
    \quad + \texttt{ShareDec} FFN $d_{\text{ff}^\prime}=4,096$ & $34$ & $4^\dag$ & $61$ & $7^\dag$ & $54$ & $1$ & $ 79$ & $7^{\pm 1.3}$ & $110$M & $(48)$  \\
    \quad + \texttt{ShareDec} FFN $d_{\text{ff}^\prime}=49,152$ & $\textbf{36}$ & $\textbf{1}$ & $\textbf{62}$ & $\textbf{9}$ & $\textbf{59}$ & $\textbf{4}$ & $69$ & $3^{\pm 0.2} $ & $202$M & $(100)$ \\
    \midrule
    Transformer Deep Enc. Shallow Dec. \len$\rightarrow$\lde & $35$ & $5$ & $\textbf{62}$ & $\textbf{4}$ & $58$ & $0$ & $ 230$ & $1^{\pm 0.8} $& $236$M & $(100)$ \\
    \quad + \texttt{ShareEncNoDec} FFN $d_{\text{ff}^\prime}=4,096$ & $34$ & $8^\dag$ & $61$ & $6^\dag$ & $55$ & $4$ & $ 235$ & $0^{\pm 0.5}$ & $127$M & $(54)$  \\
    \quad + \texttt{ShareEncNoDec} FFN $d_{\text{ff}^\prime}=57,344$ & $\textbf{35}$ & $\textbf{7}$ & $\textbf{62}$ & $\textbf{4}$ & $\textbf{58}$ & $\textbf{9}$ & $ 233$ & $5^{\pm 0.7}$ &  $236$M & $(100)$  \\
    \midrule
    Transformer Base \len$\rightarrow$\lro  & $\textbf{22}$ & $\textbf{9}$ & $\textbf{52}$ & $\textbf{9}$ & $\textbf{50}$ & $\textbf{9}$ & $119$ & $3^{\pm 1.1}$ & $64$M & $(100)$  \\
    \quad + \texttt{SharedEncNoDec} FFN $d_{\text{ff}^\prime}=2,048$ & $22$ & $2^\dag$ & $52$ & $5^\dag$ & $45$ & $8$ & $152$ & $8^{\pm 1.4}$ & $41$M & $(64)$  \\ 
    \quad + \texttt{SharedEncNoDec} FFN $d_{\text{ff}^\prime}=24,576$  & $\textbf{22}$ & $\textbf{9}$ & $52$ & $8$ & $46$ & $7$ & $150$ & $6^{\pm 0.5}$ & $64$M & $(100)$  \\
    \midrule
    Transformer Big Multilingual  &   $26$&$8$ & $46$&$3$ & $47$&$7$ & $ 94$&$6^{\pm 1.6}$ & $422$M & $(100)$ \\
    \quad + \texttt{SharedEncNoDec} FFN $d_{\text{ff}^\prime}=4,096$ & $25$&$5^\dag$ & $45$&$1^\dag$ & $40$&$8$ & $107$&$1^{\pm 1.4}$ & $330$M & $(78)$  \\
    \quad + \texttt{SharedEncNoDec} FFN $d_{\text{ff}^\prime}=49,152$  & $\textbf{28}$&$\textbf{0}^\dag$ & $\textbf{47}$&$\textbf{3}^\dag$ & $\textbf{50}$&$\textbf{7}$ & $ 111$&$5^{\pm 1.1} $ & $422$M & $(100)$  \\
    \bottomrule
    \end{tabular}
    \caption{Accuracy of One Wide FFN for \len$\rightarrow$\lde with Transformer Base, Decoder Only, and Deep Encoder Shallow Decoder on WMT$22$; for low resource \len$\rightarrow$\lro with Base version on WMT16, and multilingual with Transformer big on Flores. $\dag$ implies the system is statistical significantly different at $p < 0.05$.}
    \label{tab:full-capacity-results:b}
\end{table*}

\label{sec:beyond_ende}
So far, all our experiments focused on the Transformer Big and on WMT22 \len$\rightarrow$\lde. In this section, we apply what we learned to other architectures and language pairs. We run experiments on the low resource language direction \len$\rightarrow$\lro and a large scale multilingual model.

For \len$\rightarrow$\lde, we apply our proposal to a Transformer Base model, a Deep Encoder Shallow Decoder model \citep{Kasai0PCS21}, and a Decoder-Only model. 
For the Transformer Base, we observe an accuracy gain of $0.5$ \bleu ($2.2$ \bleu over the vanilla \texttt{SharedEncNoDec} model) and an inference speedup of around $25\%$. 
In the Deep Encoder Shallow Decoder model, we observe a more modest accuracy gain of $0.2$ \bleu points ($0.9$ \bleu over the vanilla \texttt{SharedEncNoDec} model).
However, the inference speedup from dropping the decoder FFNs is minimal ($< 1\%$), which is expected because of the small depth of the decoder in this architecture.

\paragraph{Decoder-only models} With the advent of Large Language Models (LLMs) like GPT \citep{brown2020language}, and PaLM \citep{chowdhery2022palm}, a lot of effort has been put on decoder-only Transformer models. We train a decoder-only model on WMT22 \len$\rightarrow$\lde, as shown on \Cref{tab:full-capacity-results:b}. Due to the absence of an encoder, we are limited to applying a wide FFN on the decoder side. 
As in the other setups, we get an accuracy gain of $+0.3$ \bleu over the baseline decoder-only model ($+1.7$ \bleu over \texttt{ShareDec}), but the latency degrades by $12\%$. 
This is not surprising: due to the autoregressive nature of the decoder, increasing the size of its FFN has a bigger impact on speed.

\paragraph{Low-resource languages} In \len$\rightarrow$\lro the accuracy of the One Wide FFN Model is only on par compared to the base model, even though it is a higher than the vanilla \texttt{SharedEncNoDec} model.
We hypothesize that due to the low resource condition, our proposed model already reaches saturation as there are not that many salient textual patterns to be learned by the FFN.

\paragraph{Multilingual} Finally, we observe the similar trend on the multilingual setup, where the One Wide FFN Model is $+1.2$ \spBLEU points more accurate than the baseline Transformer Big and $+2.5$ \spBLEU points more accurate than the vanilla \texttt{SharedEncNoDec},  this gain is significant in $79$ out of $90$ directions and when all tests sets are concatenated. 
Additionally, this large accuracy gain also comes with around $18\%$ inference speed-up, consistent with our previous results.




\section{Related Work}
Weight pruning and parameter sharing are well-known techniques to reduce a model's footprint. Given the scale of the latest models \citep{chowdhery2022palm}, there have been multiple efforts to prune neurons based on different automatic methods \citep{dalvi-etal-2020-analyzing,NEURIPS2019_2c601ad9,voita-etal-2019-analyzing}, sharing parameters efficiently \citep{ge-etal-2022-edgeformer,reid-etal-2021-subformer-exploring}, and factorizing certain components \citep{Lan2020ALBERT,hu2022lora}.

Neuron pruning methods often focus on finding and pruning redundant neurons through correlation methods \citep{dalvi-etal-2020-analyzing}, but also on how Transformer components like the multi-head attention can be pruned significantly due to model redundancy in the encoder or decoder either by checking the gradients salience \cite{NEURIPS2019_2c601ad9} or a differentiable relaxation of the $l_0$ regularization at training time \citep{voita-etal-2019-analyzing}.

For parameter sharing, the Universal Transformer \citep{dehghani2018universal} proposed a model where all layers are shared (\ie, in effect it reduced the model to a single shared layer).
\citet{takase-kiyono-2023-lessons} proposes finding an optimal configuration of shared layers in the encoder or decoder through different methods of sharing (in sequence, in cycle, or in reversed cycle) always keeping a specified number of final layers\footnote{See \Cref{app:custom_sharing} for a detailed description and comparison.}. Similarly, \citet{reid-etal-2021-subformer-exploring} proposes an approach where just the middle layers are shared, while the bottom and top layers are independent, and using a lower dimensionality for the embedding layer.
Analogously, \citet{ge-etal-2022-edgeformer} focus on minimizing the number of parameters and the number of calls to each parameters' group in order to optimise on-device models. They achieve this by sharing the encoder and decoder in a similar way to both previous methods, particularly by sharing all layer parameters in cycle like \citet{takase-kiyono-2023-lessons}.

Previous works also focus on reducing the dimensionality of certain parameters, mostly through low rank factorization. \citet{Lan2020ALBERT} decomposes the embedding layer into a lower rank embedding matrix and a projection to the actual hidden size while also sharing all parameters across all layers. In addition to sharing parameters efficiently, \citet{ge-etal-2022-edgeformer} proposes a lightweight decomposition of the FFN where instead of a single component there are 2 projections with a smaller dimensionality than vanilla Transformers.
Our work is close to \citet{ge-etal-2022-edgeformer} but instead of factorizing we explore sharing and full pruning of the FFN. In contrast with previous works, we also explore increasing the encoder FFN size while dropping the decoder's completely.

\section{Conclusion}
In this work, we studied the importance of the FFN in Transformer models. 
We analyzed the impact of removing and/or sharing the FFN across layers and found that, due to this component's redundancy, the model sizes can be substantially reduced with little impact on accuracy for Machine Translation. 
In particular, we found that sharing the FFN across all encoder layers while making it larger and removing it from the decoder layers leads to models that are more accurate and faster at inference.

Our findings are applicable across multiple settings, including decoder-only and multilingual models. In a low-resource setting the results are modest but our approach can still recover the baseline's performance with a faster inference.

Finally, we conducted a thorough similarity analysis between the vanilla Transformer and our proposed architectures, and found that the latter's internal representations do not differ significantly from the former's, except in that they are less redundant.

\section*{Limitations}
In this work, our focus was Machine Translation. Although we expect the results to generalize to other sequence-to-sequence tasks, further experiments are needed, which we leave for future work.

\section*{Ethics Statement}
One important consideration is the energy consumption for model training, which results in green-house emissions \citep{strubell-etal-2019-energy}.
Our work uses existing datasets, and inherits some of the risks associated with them, such as privacy leakage \citep{privacy-leakage-lm} and gender bias \citep{cho-etal-2019-measuring}. Mitigation strategies such as those from \citet{vanmassenhove-etal-2018-getting} may be necessary.

\section*{Acknowledgements}
We would like to thank Robin Schmidt, Matthias Sperber, and Stephan Peitz for their feedback and support in reviewing this work. 


\appendix

\section{Custom Sharing of Multiple FFNs}
\label{app:custom_sharing}
There is a combinatorial number of ways of sharing $M < N$ FFNs within a module of $N$ layers. Since this is prohibitive, we investigate the following strategies from \citet{takase-kiyono-2023-lessons}:

\begin{itemize}
    \item \texttt{Sequence}: assign one FFN for every $M/N$ consecutive layers, forming a block pattern.
    
    \item[] $\text{FFN}_{i}(\cdot) \overset{\text{tied}}{=} \text{FFN}_{seq_m}(\cdot), \forall i : 1 \leq i \leq N, m= \lfloor (i-1)/ (N/M) \rfloor $
    
    \item \texttt{Cycle}: stack $M$ FFNs in an identical order, forming a repetitive checkerboard pattern.
    
    \item[] $\text{FFN}_{i}(\cdot) \overset{\text{tied}}{=} \text{FFN}_{cyc_m}(\cdot), \forall i : 1 \leq i \leq N, m= (i-1) \,\,\mathbf{modulo}\,\, M $

    \item \texttt{Cycle (Rev)}: stack $M$ FFNs in a reverse order, forming a repetitive palindrome series.
    
    \item[] $\text{FFN}_{i}(\cdot) \overset{\text{tied}}{=} \text{FFN}_{cycrev_m}(\cdot), \forall i : 1 \leq i \leq N, m = N/M - i$
\end{itemize}
\noindent Note that we assume that $N$ is an even number and divisible by $N$. $\texttt{Cycle (Rev)}$ is only valid for $M=N/2$. The \texttt{EdgeFormer} \citep{ge-etal-2022-edgeformer} adopts \texttt{Cycle} with $M=2$ for the encoder FFNs.

\Cref{tab:custom_sharing} shows the results of these strategies applied on the encoder.
As references, we copy the results of the \texttt{Transformer Big} and \texttt{ShareEnc} from \Cref{tab:sharing-enc-dec-ffn}.
Not only is the accuracy of \texttt{ShareEnc} similar to \citet{takase-kiyono-2023-lessons}'s strategies, but it also uses fewer parameters and is easier to extend.

\begin{table}[ht]
    \centering
    \setlength{\tabcolsep}{1.2pt}
    \begin{tabular}{lccc}
        \toprule              
        Architecture & \bleu & $\mid \theta \mid$ & (\%) \\
        \midrule
        \texttt{Transformer Big} & $35.6$ & $228$M & ($100$) \\
        \quad + \texttt{SharedEnc (\emph{M}=1)} & $35.4$ & $186$M & ($82$) \\
        \midrule
        \quad + \texttt{Sequence \emph{M}=2} & 35.2 & 194M & (85) \\
        \quad + \texttt{Sequence \emph{M}=3} & 35.3 & 202M & (88)\\
        \quad + \texttt{Cycle \emph{M}=2} & 35.2 & 194M & (85) \\
        \quad + \texttt{Cycle \emph{M}=3} & 35.5 & 202M & (88) \\
        \quad + \texttt{Cycle Rev \emph{M}=2} & 35.2 & 194M & (85) \\
        \quad + \texttt{Cycle Rev \emph{M}=3} & 35.5 & 202M & (88) \\
        \midrule
    \end{tabular}
    \caption{Accuracy of different FFN sharing strategies on WMT22 \len $\rightarrow$ \lde.}
    \label{tab:custom_sharing}
\end{table}

\section{Sharing or Dropping Attention}
\label{app:attention}
We report the results of sharing attention modules (either self, cross or both) across layers in \Cref{tab:attention-sharing-results}. In contrast with the FFN, attention seems to play a more crucial role in the model's performance, as sharing the different attention mechanisms in both encoder and decoder causes a large accuracy drop across all settings, with the exception of sharing the decoder's cross attention and the encoder's self attention.

\begin{table}[ht!]
    \centering
    {\setlength{\tabcolsep}{1.2pt}
    \begin{tabular}{ccccr@{}r}
    \toprule              
    Encoder & \multicolumn{2}{c}{Decoder} & BLEU & \multicolumn{2}{c}{$\mid \theta \mid~(\%)$} \\
    Self-Att & Self-Att & Cross-Att & & \\
    \midrule
    \multicolumn{3}{c}{\texttt{Transformer Big}} & $35.6$ & $228\text{M}$ & $(100)$ \\
    Shared & Shared & Shared & $27.5$ & $165\text{M}$ & $(72)$ \\
    Shared & Shared & Indiv. & $27.6$ & $186\text{M}$ & $(82)$ \\
    Shared & Indiv. & Indiv. & $35.5$ & $207\text{M}$ & $(91)$ \\
    Indiv. & Shared & Indiv. & $26.5$ & $207\text{M}$ & $(91)$ \\
    Indiv. & Shared & Shared & $25.7$ & $186\text{M}$ & $(82)$ \\
    Indiv. & Indiv. &  Shared  & $35.5$ & $207\text{M}$ & $(91)$ \\
    
    \bottomrule
    \end{tabular}}
    \caption{\bleu scores on WMT 22 \len$\rightarrow$\lde when sharing the attention of both encoder and decoder (self and cross). Nomenclature follows \Cref{sec:experimental_setup} but with Self Attn an Cross Attn as the encoder/decoder's self attention and cross-attention (decoder), respectively.}
    \label{tab:attention-sharing-results}
\end{table}

\section{Details on Internal Representations Analysis}
\label{app:appendix_similarity}

\subsection{Raw Similarity Scores for Benchmarking}
\label{app:appendix_similarity:benchmark}
We establish a \textit{benchmark} score for the expected similarity of our two metrics by comparing the baseline Transformer Big with identical models trained from different random seeds. 
\Cref{tab:sharing-enc-dec-ffn-similarity-unnorm} presents the raw similarity scores from which we compute the normalized scores presented in \Cref{tab:sharing-enc-dec-ffn-similarity}.  
As shown, the similarity between 

\begin{table}[ht!]
    \centering
    {\setlength{\tabcolsep}{1.2pt}
    \begin{tabular}{lcccc}
    \toprule
    \multirow{2}*{Architecture} & \multicolumn{2}{c}{Encoder} & \multicolumn{2}{c}{Decoder} \\
    & \cka & \lns & \cka  & \lns \\
    \midrule
    \texttt{TransformerBig} Seed 2  & $.96$ & $.61$ & $.94$ & $.62$  \\
    \texttt{TransformerBig} Seed 3  & $.96$ & $.62$ & $.95$ & $.62$  \\
    \midrule
    \texttt{SharedEnc}  & \cellcolor{gray!25}$.94$ & \cellcolor{gray!25}$.58$ & $.95$ & $.62$  \\
    \texttt{SharedDec}  & $.97$ & $.62$ & \cellcolor{gray!25}$.93$ & \cellcolor{gray!25}$.59$ \\
    \texttt{SharedEncSharedDec} & \cellcolor{gray!25}$.95$ & \cellcolor{gray!25}$.59$ & \cellcolor{gray!25}$.94$ & \cellcolor{gray!25}$.59$ \\
    \texttt{SharedEncDec}  & \cellcolor{gray!25}$.94$ & \cellcolor{gray!25}$.57$ & \cellcolor{gray!25}$.93$ & \cellcolor{gray!25}$.58$ \\
    \texttt{NoEnc}  & \cellcolor{blue!25}$.87$ & \cellcolor{blue!25}$.43$ & $.95$ & $.60$  \\
    \texttt{NoDec}  & $.96$ & $.60$ & \cellcolor{blue!25}$.90$ & \cellcolor{blue!25}$.54$  \\
    \texttt{ShareEncNoDec} & \cellcolor{gray!25}$.94$ & \cellcolor{gray!25}$.59$ & \cellcolor{blue!25}$.92$ & \cellcolor{blue!25}$.55$\\
    \midrule
    \texttt{ShareEncNoDec}$^{d_\text{ff}^\prime=41952}$ & \cellcolor{gray!25}$.94$ & \cellcolor{gray!25}$.51$ & \cellcolor{blue!25}$.89$ & \cellcolor{blue!25}$.51$\\
    
    \bottomrule
    \end{tabular}}
    \caption{Raw similarity of the representations of corresponding layer-modules of different architectures vs. the Transformer Big for WMT22 \len$\rightarrow$\lde. For \textit{NoDec} configurations we compare the final output of the transformer layer as a whole as they have different sub-modules. The columns for shared and for dropped FFNs are highlighted in \colorbox{gray!25}{gray} and \colorbox{blue!25}{blue} respectively. }
    \label{tab:sharing-enc-dec-ffn-similarity-unnorm}
\end{table}

\subsection{Layer-wise Analysis}
\label{app:appendix_similarity:layerwise}
\begin{figure}[ht]
     \centering
     \begin{subfigure}[b]{0.4\textwidth}
         \centering
         \includegraphics[width=\textwidth]{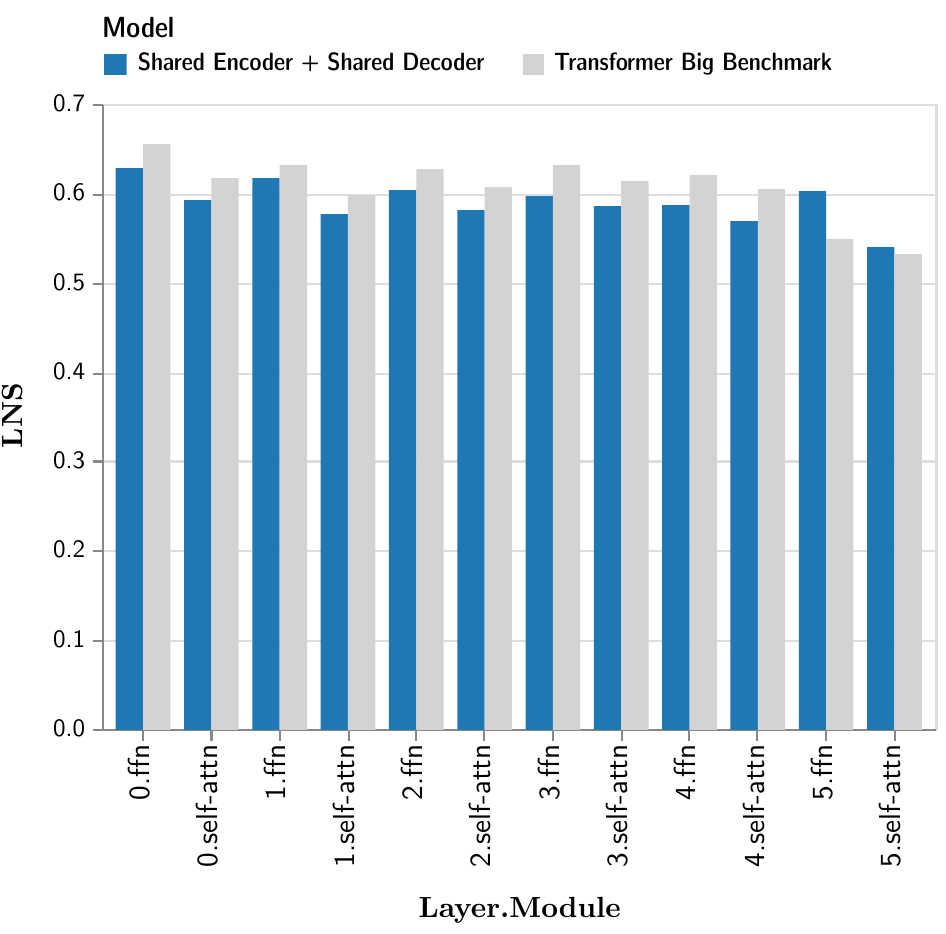}
         \caption{Encoder}
         \label{fig:layerwise_sim_shared_ffn_encoder}
     \end{subfigure}
     \hfill
     \begin{subfigure}[b]{0.4\textwidth}
         \centering
         \includegraphics[width=\textwidth]{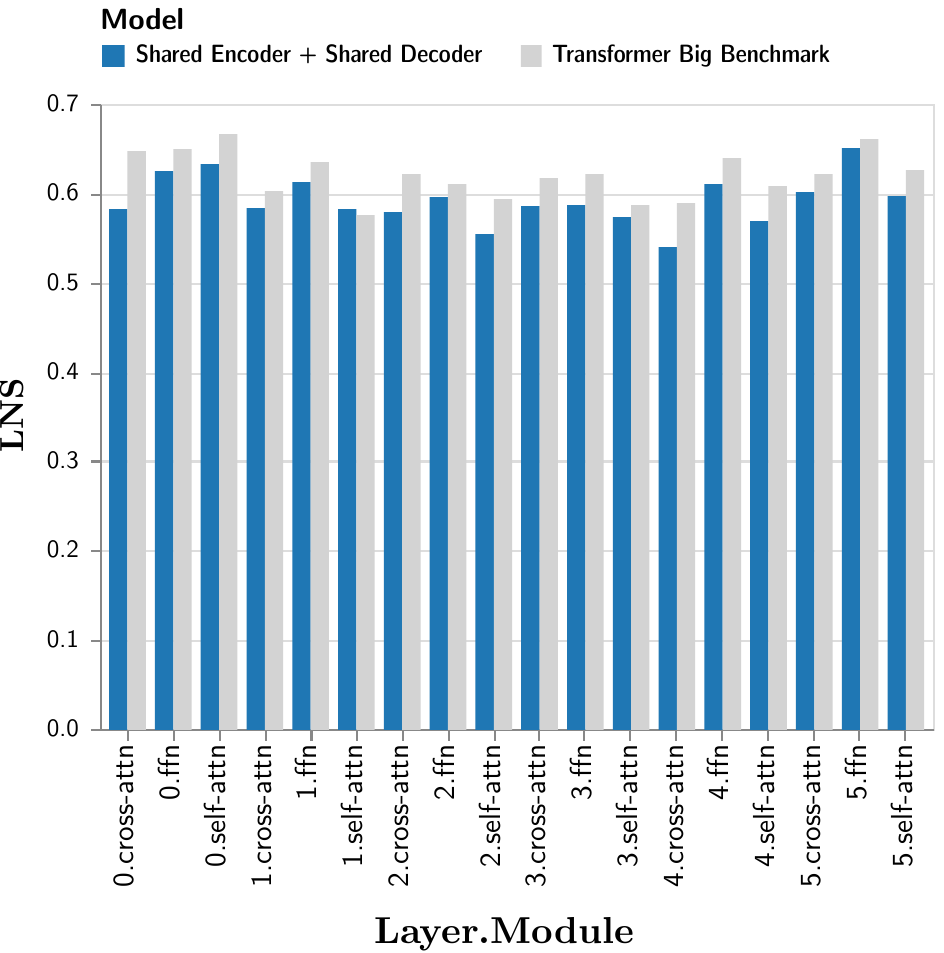}
         \caption{decoder}
         \label{fig:layerwise_sim_shared_ffn_decoder}
     \end{subfigure}
        \caption{Layerwise \lns between \texttt{SharedEncSharedDec} and \texttt{Transformer Big} (blue bars). \lns between two versions of Transformer Big trained from different random initializations are shown by the grey bars to ground the comparison. FFN sharing does not dramatically change activations produced at each layer.}
        \label{fig:layerwise_sim_shared_ffn}
\end{figure}

In \Cref{tab:sharing-enc-dec-ffn-similarity}, we report the aggregated similarity scores across all layers of Transformer encoder and decoder. 
Here, we report a more fine-grained layer-wise similarity score mostly to showcase the reliability of the aggregated scores.
In \Cref{fig:layerwise_sim_shared_ffn}, we plot layerwise \lns to study how similar the semantic information captured at each layer is to that of the baseline model at \emph{every layer}. When \lns scores are high, the network is producing similar local neighborhoods for each sentence in our evaluation set. 
In particular, we are interested in comparing the benchmark \lns scores and those of \texttt{SharedEncSharedDec} at each layer. 
As shown, the layer-wise \lns scores of \texttt{SharedEncSharedDec} track the baseline scores at almost every layer, confirming the reliability of the aggregated score.
We observe similar pattern for all the models that we evaluate in this paper.

\section{Effect of batch size on decoding speed}
\label{app:batch_size}
\begin{table*}[ht]
    \centering
    \begin{tabular}{rr@{.}lr@{.}lrr}
    \toprule              
    $\mid \text{Batch} \mid$ & \multicolumn{2}{c}{Transformer Big} & \multicolumn{2}{c}{\textit{One Wide FFN}} & $\text{Speed-up}~(\%)$ & \textit{\# batches}\\    \midrule

$1$ & $110$ & $8^{\pm 1.2}$ & $137$ & $5^{\pm1.1}$ & $24$ & $2,047$ \\
$2$ & $221$ & $7^{\pm 14.3}$ & $260$ & $9^{\pm 6.5}$ & $18$ & $1,024$ \\
$4$ & $397$ & $4^{\pm 8.0}$ & $448$ & $9^{\pm 2.0}$ & $13$ & $512$ \\
$8$ & $718$ & $3^{\pm 8.0}$ & $748$ & $7^{\pm 10.6}$ & $4$ & $256$ \\
$16$ & $1,220$ & $7^{\pm 56.2}$ & $1,226$ & $9^{\pm 17.2}$ & $1$ & $128$ \\
$32$ & $1,958$ & $5^{\pm 112.4}$ & $1,837$ & $6^{\pm 15.3}$ & $-6$ & $64$ \\
$64$ & $1,319$ & $1^{\pm36.7}$ & $1,259$ & $0^{\pm70.0}$ & $-5$ & $32$ \\
$128$ & $1,925$ & $1^{\pm64.8}$ & $1,705$ & $0^{\pm62.3}$ & $-11$ & $16$ \\
$256$ & $2,312$ & $1^{\pm67.4}$ & $1,976$ & $5^{\pm123.2}$ & $-15$ & $8$ \\
$512$ & $2,512$ & $0^{\pm50.1}$ & $1,957$ & $9^{\pm32.6}$ & $-22$ & $4$ \\

    \bottomrule
    \end{tabular}
    \caption{Effect of batch size on decoding speed (in tokens/s) for the Transformer Big and \textit{One Wide FFN} $(d_{\text{ff}^\prime}=49,152)$. $\Delta$ is the percentage change in inference speed, and \# batches is the number of batches used to evaluate. For large batch sizes, there are fewer batches (since the dataset size is fixed), which leads to higher variance in the measurements.}
    \label{tab:batch_size:a}
\end{table*}

In \Cref{sec:one-wide-ffn}, we compared the decoding speeds of the One Wide FFN model and the Transformer Big, with a batch size of 1. 
In \Cref{tab:batch_size:a}, we delve into how the decoding speed evolves as the batch size increases. 
As shown, the One Wide FFN model is faster for smaller batch sizes, but its advantage diminishes as the batch size increases, being slower than the Transformer Big for large batch sizes. We suspect this slowdown is due to the fact that the large FFN size requires higher peak memory, making the larger sizes non-optimal for this model.




\end{document}